\documentclass[conference]{IEEEtran}
\ifCLASSINFOpdf
  \usepackage[pdftex]{graphicx}
  \graphicspath{}
  \DeclareGraphicsExtensions{.pdf,.jpeg,.png,.jpg}
\else
  \usepackage[dvips]{graphicx}
  \graphicspath{}
  \DeclareGraphicsExtensions{.eps}
\fi
\usepackage{dsfont}
\usepackage{amsmath}
\ifCLASSOPTIONcompsoc
  \usepackage[caption=false,font=normalsize,labelfont=sf,textfont=sf]{subfig}
\else
  \usepackage[caption=false,font=footnotesize]{subfig}
\fi
\usepackage{color} 

\begin{document}
%
\title{Learning a Generator Model\\from Terminal Bus Data}

\author{\IEEEauthorblockN{Nikolay Stulov\IEEEauthorrefmark{1}, {Dejan J Sobajic\IEEEauthorrefmark{2}, Yury Maximov\IEEEauthorrefmark{3}\IEEEauthorrefmark{1}, Deepjyoti Deka\IEEEauthorrefmark{3}, and Michael Chertkov\IEEEauthorrefmark{3}\IEEEauthorrefmark{1}}}
\IEEEauthorblockA{\IEEEauthorrefmark{1}Skolkovo Institute of Science and Technology, Moscow, Russia}
\IEEEauthorblockA{\IEEEauthorrefmark{2} Grid Consulting LLC, San Jose, CA, USA}
\IEEEauthorblockA{\IEEEauthorrefmark{3} Theoretical Division and Center for Nonlinear Studies, Los Alamos National Laboratory, Los Alamos, NM, USA}}


%


\maketitle

\begin{abstract}
In this work we investigate approaches to reconstruct generator models from measurements available at the generator terminal bus using machine learning (ML) techniques. The goal is to develop an emulator which is trained online and is capable of fast predictive computations. The training is illustrated on synthetic data generated based on available open-source dynamical generator model. Two ML techniques were developed and tested: (a) standard vector auto-regressive (VAR) model; and (b) novel customized long short-term memory (LSTM) deep learning model. Trade-offs in reconstruction ability between computationally light but linear AR model and powerful but computationally demanding LSTM model are established and analyzed. 
\end{abstract}


%
\IEEEpeerreviewmaketitle

\section{Introduction}
Power Generator (PG) is a complex device that delivers energy to power grids. PG is described by a system of non-linear differential-algebraic equations (DAEs) of sufficiently high order, see e.g. \cite{kundur}. The generator is connected to the rest of the power system (PS) through the terminal bus. 

In this manuscript we assume that active and reactive powers ($P$ and $Q$, respectively) as well as voltage $V$ and phase $\varphi$ are observed at the terminal bus with a sufficient frequency. We then learn the generator model as a mapping between a pair of input variables $(P, Q)$ and a pair of output variables ($\varphi$, $V$). We do not assume any prior knowledge of the generator power-engineering details and thus experiment with two generic ML approaches developed for correlated linear and nonlinear time-series in an application-agnostic context. 

Our ``PG-agnostic'' approach should be contrasted with ``PG-informed'' studies, some dated back to the 1980s \cite{genlearnschulz, genlearnlee, genlearndandeno, genlearnnamba},  reconstructing parameters in the corresponding system of DAEs. Complexity of the DAEs makes the reconstruction task difficult and thus computationally expensive. In many on-line power engineering applications, where speed of prediction is primary to quality and interpretability, the PG-agnostic light and computationally efficient solutions may provide a desired 
compromise for the reconstruction task. 

To accomplish the task we have built two schemes: vector auto-regressive (VAR) scheme, which is fast and simple, yet limited due to underlying linearity assumption; and a recurrent neural network (RNN) scheme of the long short-term memory (LSTM) type, which is superior in the reconstruction quality, yet more expensive implementation-wise. 

The rest of the manuscript is organized as follows. Technical introduction, with primers on AR, LSTM and data generating procedure, is given in Section \ref{sect:preliminaries}.
Section \ref{sect:algo} describes in sufficient detail our VAR and LSTM algorithmic solutions.  
Finally, data generation and learning experiments are described and analyzed in Section \ref{sect:results}. 

\section{Technical Introduction}\label{sect:preliminaries}
In this section we revisit classical algorithms for time series prediction and review our custom NN LSTM solutions for the problem. 

\subsection{Auto-regressive process}\label{subsect:ar}
Auto-regressive (AR) model is a proven tool in many stationary time series applications.  
In its multivariate version AR model becomes vector auto-regressive model (VAR) \cite{multivariative} described by the following equation
\begin{equation}
    \mathbf{y}_t = \mathbf{\mu} + \sum\limits_{i=1}^p \mathbf{A}_i \mathbf{y}_{t-i} + \mathbf{\varepsilon}_t,
    \label{eq:VAR}
\end{equation}
where $\mathbf{y}_t$ is the time series, $\mathbf{\mu}$ and $\mathbf{A}_i$ are the parameters to be learned, $\mathbf{\varepsilon}_t$ is a stochastic white noise process and $p$ is the order of the VAR model.

When the series depend on some other exogenous, changing in time characteristics, $\mathbf{x}_t$, the model becomes
\begin{equation}
    \mathbf{y}_t = \mathbf{\mu} + \sum\limits_{i=1}^p \mathbf{A}_i \mathbf{y}_{t-i} + \mathbf{\beta} \sum\limits_{i=1}^p \mathbf{A}_i \mathbf{x}_{t-i} + \mathbf{\varepsilon}_t, 
    \label{eq:VAR-exogeneous}
\end{equation}
where $\mathbf{\beta}$ is an additional parameter which needs to be learned.

\subsection{Long Short-Term Memory Network}\label{subsect:lstm}
Fully-connected neural networks (NNs) are powerful state-of-the-art tools for many ML tasks. In order to improve NNs' task-specific generalization ability new architectures are proposed. For example, convolutional neural networks (CNNs) were designed for images \cite{krizhevsky2012imagenet}. CNNs take advantage of the image (postulated) translational invariance,  thus resulting in a significant reduction in the number of training parameters. In the case of a time-stationary data sequence similar reduction in the number of training parameters is achieved through the so-called Recurrent Neural Networks (RNNs) proposed in \cite{rnn}, characterized by time-invariant (or quasi-time invariant,  i.e. changing slowly with time) couplings between the signal at different time-frames. However, earlier tests of RNN have also revealed a number of significant drawbacks in performance. The long short-term memory (LSTM)~\cite{lstm} NN extending RNNs with the concept of memory cell were designed to overcome such issues. Memory cell enables RNN to learn when to remember and when to forget previously read data. As a result, LSTMs offer better performance in the task of modelling long time series.

\subsection{Synthetic Data for the Ground Truth}\label{subsec:groundtruth}

\begin{figure}[!h]
    \centering
    \includegraphics[width=\linewidth]{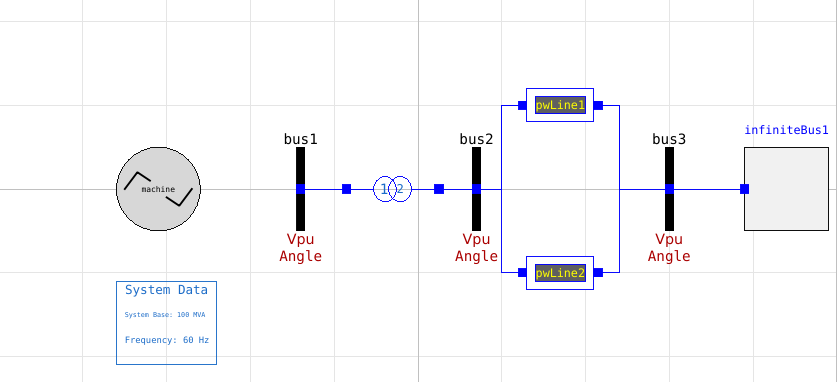}
    \includegraphics[width=\linewidth]{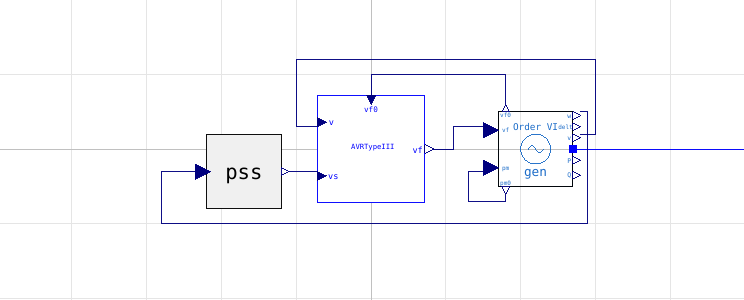}
    \caption{Example grid (top) and machine (bottom) from \cite{kundur} (example 13.2).}
    \label{fig:grid+machine}
\end{figure}

In this work we choose to work with synthetically generated data. The data was generated for a typical power generator with OpenModelica using openly distributed package OpenIPSL \cite{ipsl}. Specifically,  we use OpenIPSL implementation of a generator in a simple grid example 13.2 from~\cite{kundur}, illustrated in~Fig.~\ref{fig:grid+machine}. 
The model is initiated with static power flow (PF) values calculated with PSAT \cite{psat}. To imitate uncertainty, exogenous to the generator, we add stationary perturbations to the model in the form of stochastic faults described in the following. \footnote{Notice that this choice of uncertainty is due to the fact that other characteristics of the power grid are hard-coded in the version of OpenModelica and OpenIPSL available to us, not allowing to inject into the model endogenously varying characteristics in any other way then generating a sequence of faults.}
The faults are introduced at the connection between the second bus and the first power line in Fig.~\ref{fig:grid+machine}. In our experiments we set fault resistance to $0.01$ pu and select fault reactance such that the generator survives the fault (does not disconnect), while showing some  visible non-linearity. We later alter these parameters to test our models in different regimes (linear and nonlinear) and fine-tune their performance and robustness. We find out, that resistance value of $10^{-5}$ is the smallest which can be stabilized by the generator (also equipped with the Power System Stabilizer). We also alter the parameters of the generator randomly by sampling values from the normal distribution, $\mathcal{N}(v, 0.1 v)$, where $v$ is the static value used in the previous experiments.

Dynamic stochasticity is introduced into the model by scheduling faults according to the following pair of master equations describing statistics of a stochastic stationary telegraph process
\begin{equation}
    \begin{cases}
        \frac{\partial}{\partial t}\mathds{P}(a, t \, | \, x, t_0) &= - \lambda \mathds{P}(a, t \, | \, x, t_0) + \mu \mathds{P}(b, t \, | \, x, t_0)\\
        \frac{\partial}{\partial t}\mathds{P}(b, t \, | \, x, t_0) &= \lambda \mathds{P}(a, t \, | \, x, t_0) - \mu \mathds{P}(b, t \, | \, x, t_0).
    \end{cases}
    \label{eq:telegraph}
\end{equation}
Note that solution of Eq.~(\ref{eq:telegraph}) initialized with, $P(a,0)=0$ and $P(b,0)=1$ (no fault at $t=0$), becomes
\begin{equation}
    \begin{cases}
        \mathds{P}(a, t \, | \, \pi) &= \frac{\mu}{\lambda + \mu} + \frac{\lambda \pi_a - \mu \pi_b}{\lambda + \mu} e^{-(\lambda + \mu) t}\\
    \mathds{P}(b, t \, | \, \pi) &= \frac{\lambda}{\lambda + \mu} + \frac{- \lambda \pi_a + \mu \pi_b}{\lambda + \mu} e^{-(\lambda + \mu) t},
    \end{cases}
\end{equation}
where the first and second terms in the sum correspond, respectively, to the stationary values, and to the exponentially decreasing with time finite memory correction. Eqs.~(\ref{eq:telegraph}) are included (coded) explicitly in the computational scheme so that at each time step, the fault state is sampled from the resulting distribution. It is straightforward to check that the  result of this scheme is the generation of a Bernoulli distributed random variable with success probability $\mathds{P}(a, t \, | \, \pi)$. In addition we also randomize at each step the resistance and reactance of the fault.

\begin{figure}[!t]
    \centering
    \subfloat[Process]{\includegraphics[width=1.7in]{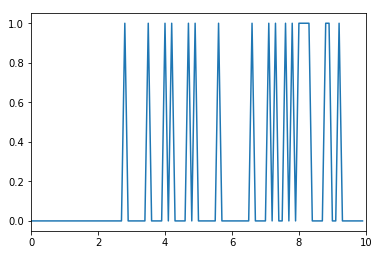}%
    \label{fig:telegraph:proc}}
    \hfil
    \subfloat[Autocorrelation]{\includegraphics[width=1.7in]{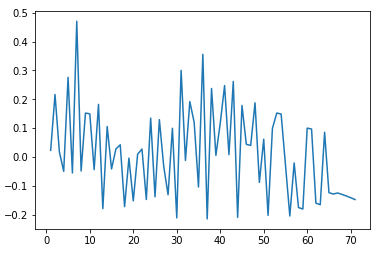}%
    \label{fig:telegraph:corr}}
    \caption{Telegraph process.}
    \label{fig:telegraph}
\end{figure}

We set $\mu = 0.03$ and $\lambda = 0.02$ to bias towards an open fault and to randomize the time of the first fault occurrence.  An examplary process and its auto-correlation (as a function of the time lag) are shown in Fig.~\ref{fig:telegraph:proc} and Fig.~\ref{fig:telegraph:corr} respectively. The auto-correlation is computed as Pearson correlation, Eq.~\eqref{eq:corr}, between $x$ at the observation time and $x$ at the time shifted by a lag. Observe that the auto-correlation resembles white noise, which indicates that the process is almost memoryless.

\begin{equation}
    r_{xy} = \frac{\sum_{i=1}^T (x_i - \overline{x})(y_i - \overline{y})}{\sqrt{\sum_{i=1}^T (x_i - \overline{x})^2}\sqrt{\sum_{i=1}^T (y_i - \overline{y})^2}}
    \label{eq:corr}
\end{equation}

The OpenIPSL solver requires time resolution to be at least $10^5$ steps per second. On the other hand standard PMU measurements are recorded $10-50$ times per second. These considerations motivate us to analyze the the practical case of 10 measurements per second.

\section{Algorithms, Diagnostic and Performance}\label{sect:algo}

\subsection{Vector AR}\label{subsect:var}
\begin{figure*}[ht!]
    \centering
    \includegraphics[width=\linewidth]{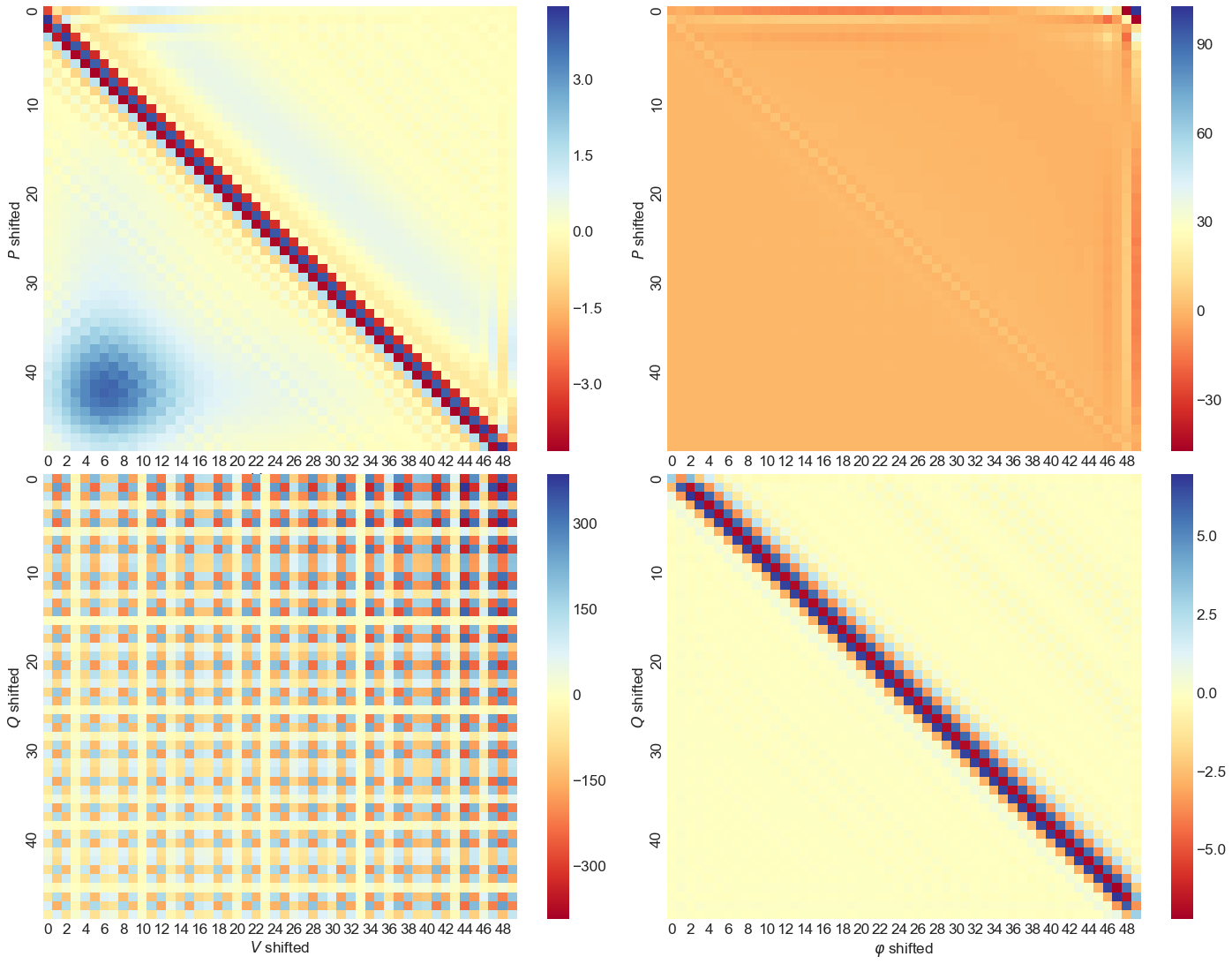}
    \caption{Colormap interpretation of inverted correlation matrices for each pair of variables. Upper triangular part of each plot corresponds to degree of conditional dependence of first variable's present and second variable's past. Largest sequence of non-zero elements ending with zeros constitutes a measure for model order $p$.}
    \label{fig:correlations}
\end{figure*}

In order to determine the optimal order of VAR model we carry out several experiments computing correlations between input and output variables. For each pair of vector-variables (two time-series) we construct a matrix of correlations, $C$, between the two variables shifted by different time lags (columns for lags in first variable and rows for lags in the second). Next we invert $C$ and investigate non-zero elements. For Gaussian noise, non-zero elements of the upper triangle of $C^{-1}$ signifies conditional independence (lack of correlations) between first variable's present and second variable's past, while non-zero elements of the lower triangle denotes that the present value of the second variable is conditionally independent of the first variable's past \cite{wainwright2008graphical}. The resulting $C^{-1}$ matrices for each pair of variables are shown in the form of a heat-map in Fig.~\ref{fig:correlations}.

We derive from the heat maps in Fig.~\ref{fig:correlations}, that for all pairs of variables, but the active power $P$ and phase $\varphi$, the length of strip of non-zero elements does not exceed 5 time steps. This observation means that after~5 time steps all variables except $P$ and $\varphi$ become conditionally independent. For the case of $P$ and $\varphi$ we observe notable dependence with lags of up to~32. We conclude that this pair of variables is conditionally independent after 32 time steps. Based on this derivation, we set the order of VAR model to be equal to~32.

\begin{figure}[h!]
    \centering
    \includegraphics[width=\linewidth]{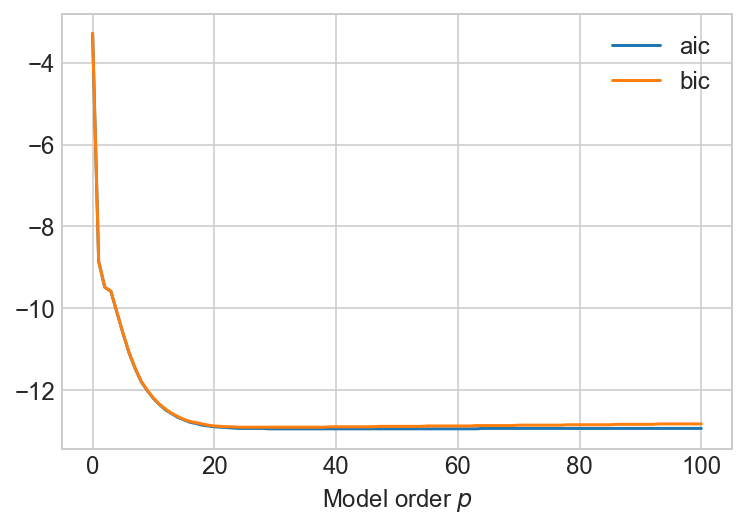}
    \caption{Comparison of AIC and BIC.}
    \label{fig:errsbest1}
\end{figure}

This decision is further supported by implementation-specific information criteria-based order selection methods. We have experimented with multiple criterion, including Akaike information criterion (AIC, see Eq.~\ref{eq:akaike}) \cite{akaike}, final prediction error (FPE) \cite{akaike}, Hannan-Quinn information criteria (HQIC) \cite{claeskens2008model} and Bayesian information criterion (BIC) \cite{claeskens2008model}. 
We find, that AIC yields the most stable results for our case. 
\begin{equation}
    \mathrm{AIC} = 2k - 2 \ln \mathcal{L}
    \label{eq:akaike}
\end{equation}
where $k$ is the number of parameters and $\mathcal{L}$ is the maximum likelihood achieved by the model. It reports the same 32 time steps lag as what we have concluded from Fig.~\ref{fig:errsbest1}

Notice that the input variables, e.g. voltage and phase, are per se not statistically stationary. We observe, however, that the assumption of statistical stationarity applies much better to the characteristics' temporal derivatives (or equivalently difference/increment between the values in two adjacent time slots). Therefore, we apply VAR to the temporal increments. The LSTM model, in contrast, does not require statistical stationarity, which explains why we apply it directly to the input variables.

\subsection{Weight-Dropped (WD) LSTM}\label{subsect:wdlstm}
To boost performance on the low-resolution data, we follow the state of the art in the LSTM modeling \cite{awdlstm} and use the Weight-Dropped LSTM. Two other specifics of the scheme we use are related to the use of the multiple LSTM approach and its applications to the supervised setting (input-output relation). The multiple LSTM architecture, utilizing a number of different training techniques, e.g. dropout, hidden-to-hidden dropout, weight decay and varied sequence length, is considered empirically superior to other ML options because of its ability to capture wide and varying ranges of data dependencies.  
We use \cite{ulmfit} for implementation of the supervised scheme. 

\section{Experiments and results}\label{sect:results}

\subsection{VAR model experimental setup}
Based on the preliminary analysis of the data, we train a vector autoregressive model of order 32 with the input vector $(P, Q)$ and the output vector of the ($\varphi$, $V$) time increments.

Generation of the input-output training data is described in Section \ref{subsec:groundtruth}. 

Since the VAR model is purely linear, it is important to investigate its dependence on the change in the noise magnitude. In order to do that, we construct  4 pairs of series with fault reactances set to zero and fault resistances set to, 10 pu, 1 pu, 0.1 pu and 0.01 pu respectively. We then train 4 models on first series of each pair and test them on the second series of the corresponding pair.

\subsection{LSTM model experimental setup}

We train the neural network to read $(V, \varphi)$ pairs sequentially and output $(P, Q)$ in a (varied-length) window. The objective functional is the mean squared error (MSE) between model outputs at each step and the ground truth data.

We train the model on 200 samples with fault parameters as described above. Then one evaluates the model on 200 samples with different and additionally randomized fault parameters. Finally, we perform further testing with randomized generator parameters and extreme noise, as discussed later. 

\subsection{Estimation of quality}
To measure the algorithm quality we use normalized root mean squared error (NRMSE), which allows us to treat error as the percentage described by
\begin{equation}
    NRMSE(x, y) =  \frac{\sqrt{\frac{1}{T}\sum\limits_{t=0}^T || x_t - y_t ||_2}}{\sqrt{\frac{1}{T} \sum\limits_{t=0}^T || x_i ||_2}},
\end{equation}
where $x$ is the ground truth, $y$ is the estimation and $|| \cdot ||_2$ is the notation for  the $\ell_2$ norm.

\subsection{Results}
Quantitative comparison of all the models tested is summarized in Table~\ref{tab:quality}.

\subsubsection{VAR}
We begin with training and testing this model on our most complicated data with randomized fault and generator parameters, denoted \textsc{randomized} in Table~\ref{tab:quality}. The model performs surprisingly well. It learns different generator and noise parameters.

\begin{figure}[ht]
    \centering
    \includegraphics[width=\linewidth]{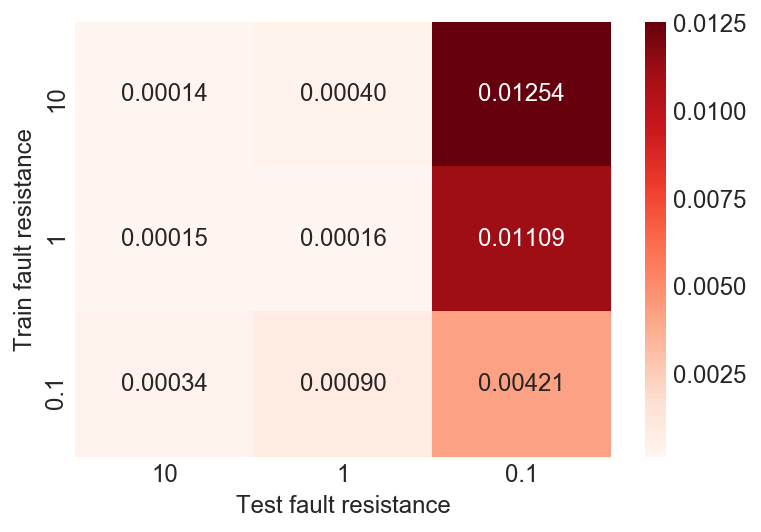}
    \caption{Color map of the VAR performance.}
    \label{fig:varcolormap}
\end{figure}

\begin{figure}[h!]
    \centering
    \includegraphics[width=.8\linewidth]{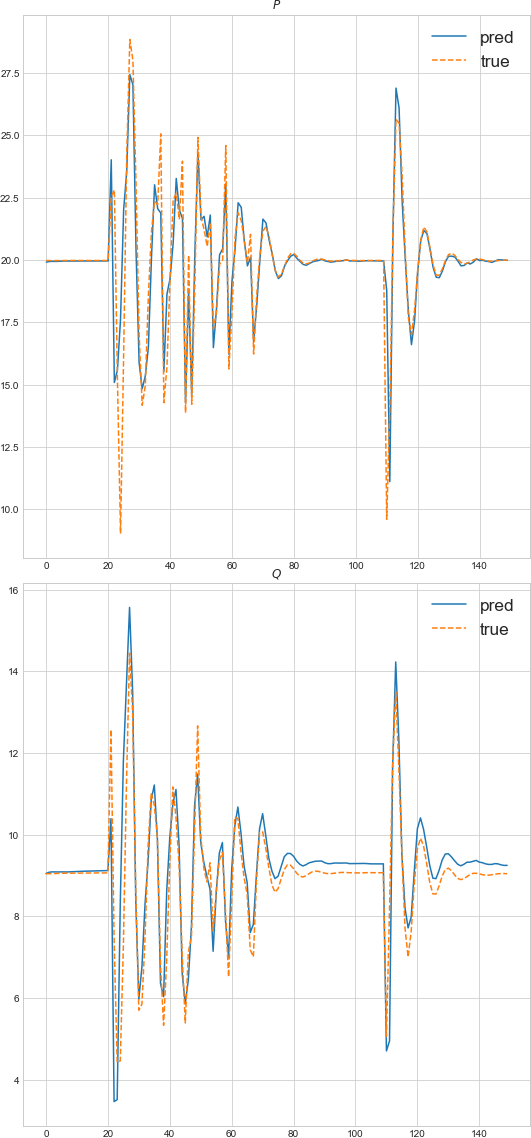}
    \caption{VAR: predicted values vs true values.}
    \label{fig:varvalstst}
\end{figure}

The prediction quality degrades with time and reactive power ($Q$) estimation is generally worse (see Fig.~\ref{fig:varvalstst}). There are ways to handle the first issue, but the second one requires an additional investigation.
\begin{table}[!h]
\renewcommand{\arraystretch}{1.3}
\caption{NRMSE statistics (in \%)}
\label{tab:quality}
\centering
\begin{tabular}{|c|c|c|c|c|}
\hline
Model & Data & Mean & Median & 95\% percentile \\ \hline
VAR & randomized & 4.72 & 4.59 & 6.52 \\ \hline
WD-LSTM & regular & 0.23 & 0.08 & 0.91 \\ \hline
WD-LSTM & randomized & 7.48 & 7.30 & 9.22 \\ \hline
FT-WD-LSTM & randomized & 0.69 & 0.60 & 1.37 \\ \hline
FT-WD-LSTM & high-order noise & 0.66 & 0.58 & 1.20 \\ \hline
\end{tabular}
\end{table}
As already stated above, we find this model noteworthy due to its ability to sense, catch and benchmark non-linearities in the data. The results of the analysis are also shown in  Fig.~\ref{fig:varcolormap} in the form of a color map.

\begin{figure}[ht]
    \centering
    \includegraphics[width=\linewidth]{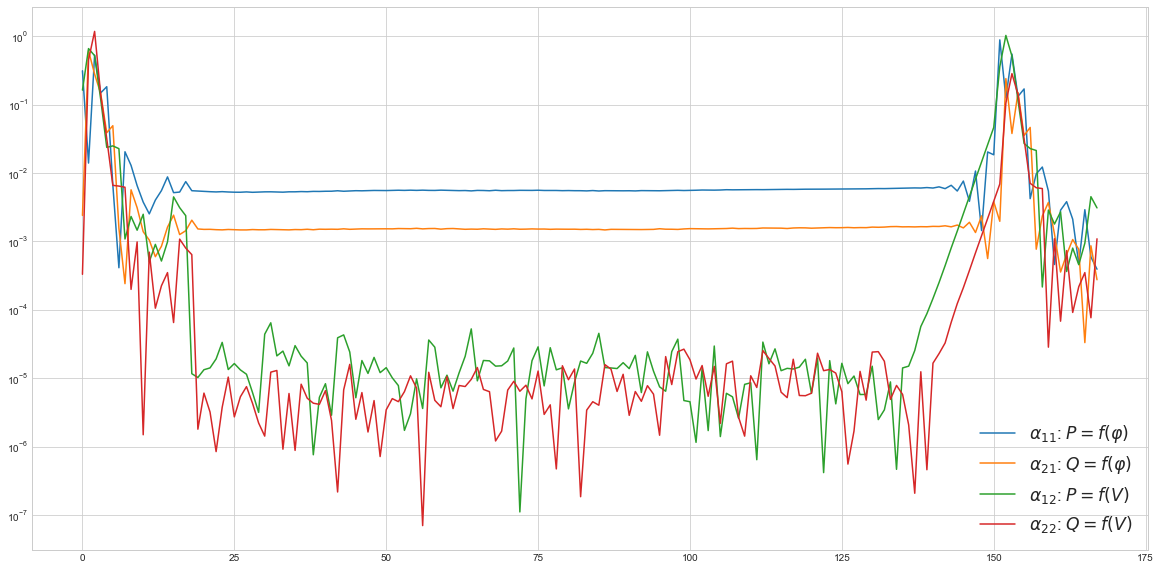}
    \caption{Decay of the learned parameter for VAR - principal scheme to reconstruct the optimal order of the scheme from the data.}
    \label{fig:varparamsdec}
\end{figure}

Finally, we investigate relation between magnitude of the noise magnitude and the order of the model, $p$. Direct application of Akaike criterion yields a counter-intuitive result that the optimal model order $p$ increases when noise magnitude becomes smaller. We relate it to overfiting, and suggest a different method to identify the optimal order. It seems more appropriate to base the cruterium on the following how the learned parameter, $A_i$, saturates with $i$: $p = \arg\min_i \{\|A_i - A_{i-1}\| < \epsilon\}$. For majority of  cases the optimal order of the scheme derived this way returns a sensible result, see  Fig.~\ref{fig:varparamsdec}.

\subsubsection{WD-LSTM}
\begin{figure}[h!]
    \centering
    \includegraphics[width=\linewidth]{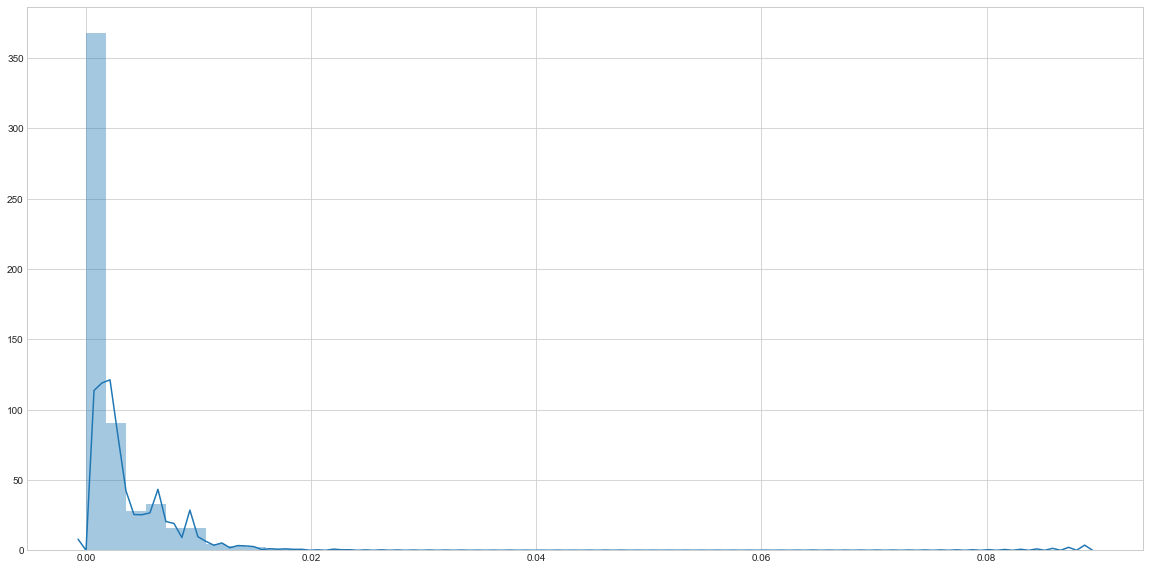}
    \caption{WD-LSTM: Estimated density of the NRMSE distribution.}
    \label{fig:errsbest}
\end{figure}

\begin{figure}[t]
    \centering
    \includegraphics[width=.85\linewidth]{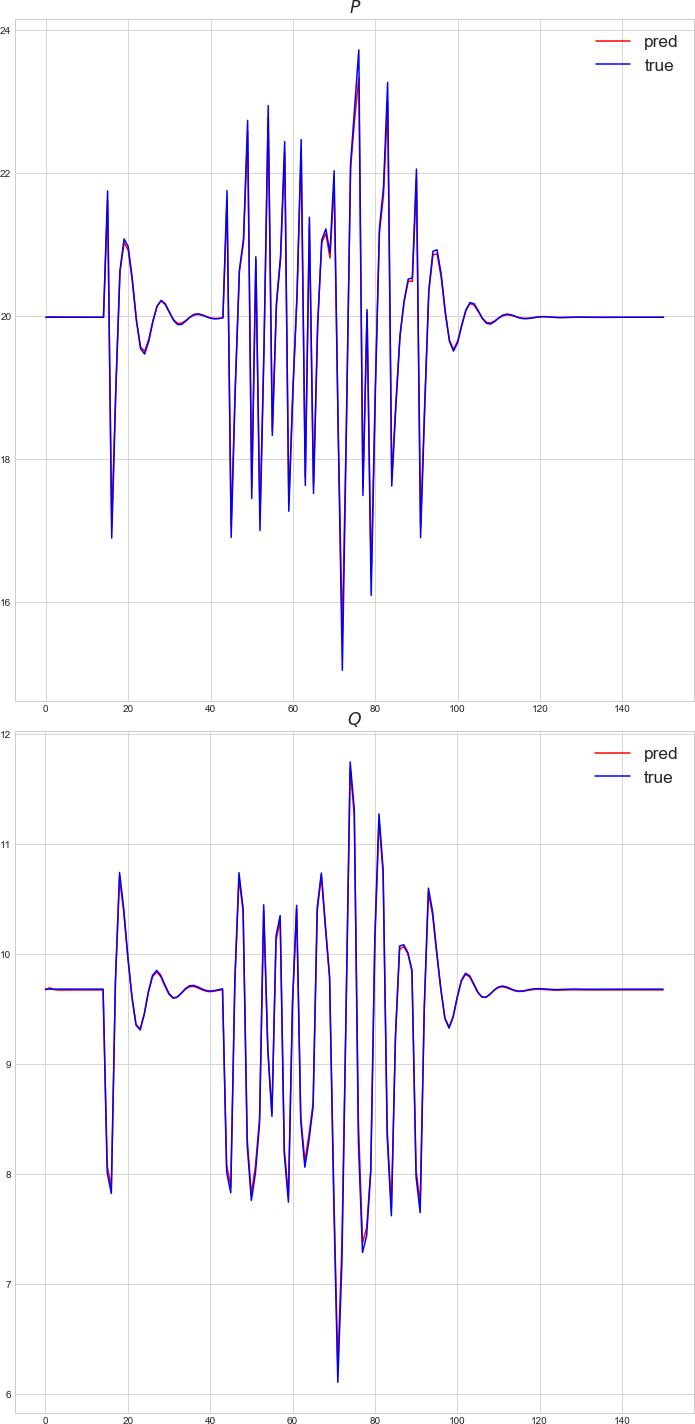}
    \caption{WD-LSTM: predicted values vs True values.}
    \label{fig:valststbest}
\end{figure}

Let us first process \textsc{normal-operation} data correspondent to fixed generator parameters. Reconstructed and actual results shown in Fig.~\ref{fig:valststbest} are almost indistinguishable. The error distribution among samples in Fig.~\ref{fig:errsbest} is also skewed to zero.

Given that the model shows such a a good performance in the simple normal case, it is important to furhter experiment putting the model in a significantly more adversarial (towards learning) regime characterized by seious fluctuations.  We train the model on the data correspondent to an additional (artificial) randomization in some parameters of the generator. This modification, as shown in Table~\ref{tab:quality}, allows our model to achieve quality comparable to the one provided by the original WD-LSTM model. We conclude that  this architecture allows generalizations over the range of different (but reasonably close) generator parameters.

Further tests shown in Table~\ref{tab:quality} include learning in the regime of the \textsc{high-order noise} data with extremely high perturbations, but fixed (not evolving) generator parameters. Resistance of the faults is set to be three orders less than for training (this is the maximum order the generator stabilizer can handle).

\clearpage

\section{Conclusions and Path Forward}\label{sect:conclusion}
%
%
In this work we have introduced a novel method, which is light and accurate for learning the model of a power generator. We have also design our custom synthetic data generation process to provide the data needed to train the models. We show that the LSTM method is truly non-linear and robust to high-order perturbations and randomization of the generator parameters, while the other, linear regression, is light in implementation but limited to regimes with small nonlinearities. We believe that such models are worth incorporating in the current grid engineering practice. 

In terms of the path forward,  we plan 
\begin{itemize}
\item extending these two already calibrated approaches by a principally different one utilizing physical generator model(s);
\item testing the devised algorithms on real (and not only synthetic) data;
\item extend the approach to modeling power distributions, as seen aggregated on the transmission level;
\item extend the approach to modeling (learning silent malfunctions as they occur) in various other precious assets of the power systems, e.g. transformers. 
\item integrated the developed schemes into a tool box monitoring larger power systems, including multiple generators, loads and transformers. 
\end{itemize}

\section*{Acknowledgment}
The work at LANL was carried out under the auspices of the National Nuclear Security Administration of the U.S. Department of Energy under Contract No. DE-AC52-06NA25396. The work was partially supported by DOE/OE/GMLC and LANL/LDRD/CNLS projects.

\bibliographystyle{IEEEtran}

\bibliography{paper.bib}

\begin{thebibliography}{10}
\providecommand{\url}[1]{#1}
\csname url@samestyle\endcsname
\providecommand{\newblock}{\relax}
\providecommand{\bibinfo}[2]{#2}
\providecommand{\BIBentrySTDinterwordspacing}{\spaceskip=0pt\relax}
\providecommand{\BIBentryALTinterwordstretchfactor}{4}
\providecommand{\BIBentryALTinterwordspacing}{\spaceskip=\fontdimen2\font plus
\BIBentryALTinterwordstretchfactor\fontdimen3\font minus
  \fontdimen4\font\relax}
\providecommand{\BIBforeignlanguage}[2]{{%
\expandafter\ifx\csname l@#1\endcsname\relax
\typeout{** WARNING: IEEEtran.bst: No hyphenation pattern has been}%
\typeout{** loaded for the language `#1'. Using the pattern for}%
\typeout{** the default language instead.}%
\else
\language=\csname l@#1\endcsname
\fi
#2}}
\providecommand{\BIBdecl}{\relax}
\BIBdecl

\bibitem{kundur}
P.~Kundur, N.~Balu, and M.~Lauby, \emph{Power system stability and control},
  ser. EPRI power system engineering series.\hskip 1em plus 0.5em minus
  0.4em\relax McGraw-Hill, 1994.

\bibitem{genlearnschulz}
R.~P. Schulz, C.~J. Goering, R.~G. Farmer, S.~M. Bennett, D.~A. Selin, and
  D.~K. Sharma, ``Benefit assessment of finite-element based generator
  saturation model,'' \emph{IEEE Transactions on Power Systems}, vol.~2, no.~4,
  pp. 1027--1033, Nov 1987.

\bibitem{genlearnlee}
C.~Lee and O.~T.~Tan, ``A weighted-least-squares parameter estimator for
  synchronous machines,'' \emph{Power Apparatus and Systems, IEEE Transactions
  on}, vol.~96, pp. 97 -- 101, 02 1977.

\bibitem{genlearndandeno}
P.~L. Dandeno, P.~Kundur, A.~T. Poray, and H.~Z. El-din, ``Adaptation and
  validation of turbogenerator model parameters through on-line frequency
  response measurements,'' \emph{IEEE Transactions on Power Apparatus and
  Systems}, vol. PAS-100, no.~4, pp. 1656--1664, April 1981.

\bibitem{genlearnnamba}
M.~Namba, T.~Nishiwaki, S.~Yokokawa, and K.~Ohtsuka, ``Idenntification of
  parameters for power system stability analysis using kalman filter,''
  \emph{IEEE Transactions on Power Apparatus and Systems}, vol. PAS-100, no.~7,
  pp. 3304--3311, July 1981.

\bibitem{multivariative}
H.~L{\"u}tkepohl, \emph{New introduction to multiple time series
  analysis}.\hskip 1em plus 0.5em minus 0.4em\relax Springer Science \&
  Business Media, 2005.

\bibitem{krizhevsky2012imagenet}
A.~Krizhevsky, I.~Sutskever, and G.~E. Hinton, ``Imagenet classification with
  deep convolutional neural networks,'' in \emph{Advances in neural information
  processing systems}, 2012, pp. 1097--1105.

\bibitem{rnn}
J.~Schmidhuber, ``The neural bucket brigade,'' in \emph{Connectionism in
  Perspective}, R.~Pfeifer, Z.~Schreter, Z.~Fogelman, and L.~Steels, Eds.\hskip
  1em plus 0.5em minus 0.4em\relax Amsterdam: Elsevier, North-Holland, 1989,
  pp. 439--446.

\bibitem{lstm}
S.~Hochreiter and J.~Schmidhuber, ``Long short-term memory,'' \emph{Neural
  Computation}, vol.~9, no.~8, pp. 1735--1780, 1997.

\bibitem{ipsl}
L.~Vanfretti, T.~Rabuzin, M.~Baudette, and M.~Murad, ``itesla power systems
  library (i{PSL}): A modelica library for phasor time-domain simulations,''
  \emph{SoftwareX}, vol.~5, pp. 84 -- 88, 2016.

\bibitem{psat}
F.~Milano, ``An open source power system analysis toolbox,'' \emph{IEEE
  Transactions on Power Systems}, vol.~20, no.~3, pp. 1199--1206, Aug 2005.

\bibitem{wainwright2008graphical}
M.~J. Wainwright and M.~I. Jordan, ``Graphical models, exponential families,
  and variational inference,'' \emph{Foundations and Trends{\textregistered} in
  Machine Learning}, vol.~1, no. 1--2, pp. 1--305, 2008.

\bibitem{akaike}
H.~Akaike, ``Information theory and an extension of the maximum likelihood
  principle,'' in \emph{Selected papers of {H}irotugu {A}kaike}.\hskip 1em plus
  0.5em minus 0.4em\relax Springer, 1998.

\bibitem{claeskens2008model}
G.~Claeskens and N.~L. Hjort, ``Model selection and model averaging,''
  \emph{Cambridge Books}, 2008.

\bibitem{awdlstm}
\BIBentryALTinterwordspacing
S.~Merity, N.~S. Keskar, and R.~Socher, ``Regularizing and optimizing {LSTM}
  language models,'' \emph{CoRR}, vol. abs/1708.02182, 2017. [Online].
  Available: \url{http://arxiv.org/abs/1708.02182}
\BIBentrySTDinterwordspacing

\bibitem{ulmfit}
\BIBentryALTinterwordspacing
J.~Howard and S.~Ruder, ``Fine-tuned language models for text classification,''
  \emph{CoRR}, vol. abs/1801.06146, 2018. [Online]. Available:
  \url{http://arxiv.org/abs/1801.06146}
\BIBentrySTDinterwordspacing

\end{thebibliography}

\end{document}